\documentclass[10pt,twocolumn,letterpaper]{article}

\usepackage{cvpr}
\usepackage{times}
\usepackage{epsfig}
\usepackage{graphicx}
\usepackage{amsmath}
\usepackage{amssymb}
\usepackage{multirow}
\usepackage{float}
\usepackage{times}
\usepackage{tabularx}
\usepackage{paralist}
\usepackage{cite}
\usepackage[nopar]{lipsum}
\usepackage{diagbox} 
\usepackage{cancel}
\usepackage{subfigure}

\usepackage{paralist}

\usepackage[pagebackref=true,breaklinks=true,letterpaper=true,colorlinks,bookmarks=false]{hyperref}

\usepackage{xspace}

\newcommand{\inEye}{\ensuremath{\textbf{\textit{I}}_{i}}\xspace}

\newcommand{\transEye}{\ensuremath{\textbf{\textit{I}}^{t}_{i}}\xspace}
\newcommand{\outEye}{\ensuremath{\textbf{\textit{I}}_{o}}\xspace}
\newcommand{\redEye}{\ensuremath{\textbf{\textit{I}}_{red}}\xspace}

\newcommand{\idWarp}{\ensuremath{\textbf{\textit{w}}}_{\ensuremath{id}}\xspace}
\newcommand{\Warp}{\ensuremath{\textbf{\textit{w}}}\xspace}
\newcommand{\xWarp}{\ensuremath{\textbf{\textit{w}}_{\ensuremath{h}}}\xspace}
\newcommand{\yWarp}{\ensuremath{\textbf{\textit{w}}_{\ensuremath{v}}}\xspace}

\newcommand{\GNetInfer}{\ensuremath{\mathcal{G}}\xspace}

\newcommand{\RNet}{\ensuremath{\mathcal{R}_{\theta}}\xspace}
\newcommand{\RSubNet}{\ensuremath{f_{\mathcal{R}_{\theta}}}\xspace}
\newcommand{\RSubNetDecode}{\ensuremath{f_{\mathcal{R}_{\theta^{*}}}}\xspace}
\newcommand{\RSubNetInfer}{\ensuremath{f_{\mathcal{R}}}\xspace}
\newcommand{\GNet}{\ensuremath{\mathcal{G}_{\phi}}\xspace}
\newcommand{\LNet}{\ensuremath{\mathcal{A}_{\psi}}\xspace}
\newcommand{\LSubNet}{\ensuremath{f_{\mathcal{A}_{\psi}}}\xspace}

\newcommand{\gGen}{\ensuremath{\mathcal{W}}\xspace}

\newcommand{\deltaV}{\ensuremath{\Delta \textit{r}}\xspace}

\newcommand{\rep}{\ensuremath{\textit{r}}\xspace}
\newcommand{\repP}{\ensuremath{\textit{r}_p}\xspace}
\newcommand{\repY}{\ensuremath{\textit{r}_y}\xspace}
\newcommand{\deltaRep}{\ensuremath{\Delta \textit{r}}\xspace}

\newcommand{\deltaGazeP}{\ensuremath{\Delta \textit{r}_p}\xspace}
\newcommand{\deltaGazeY}{\ensuremath{\Delta \textit{r}_y}\xspace}

\newcommand{\GazeP}{\ensuremath{\textit{g}_p}\xspace}
\newcommand{\GazeY}{\ensuremath{\textit{g}_y}\xspace}

\newcommand{\kP}{\ensuremath{\textit{k}_p}\xspace}
\newcommand{\kY}{\ensuremath{\textit{k}_y}\xspace}

\newcommand{\bP}{\ensuremath{\textit{b}_p}\xspace}
\newcommand{\bY}{\ensuremath{\textit{b}_y}\xspace}

\newcommand{\kNet}{\ensuremath{\textit{\textbf{k}}^{\phi}}\xspace}
\newcommand{\kPNet}{\ensuremath{\textit{\textbf{k}}^{\phi}_p}\xspace}
\newcommand{\kYNet}{\ensuremath{\textit{\textbf{k}}^{\phi}_y}\xspace}

\newcommand{\bNet}{\ensuremath{\textit{\textbf{b}}^{\phi}}\xspace}
\newcommand{\bPNet}{\ensuremath{\textit{\textbf{b}}^{\phi}_p}\xspace}
\newcommand{\bYNet}{\ensuremath{\textit{\textbf{b}}^{\phi}_y}\xspace}

\newcommand{\featureFC}{\ensuremath{\textit{\textbf{x}}}\xspace}

\newcommand{\pixelLoss}{\ensuremath{\mathcal{L}_{p}}\xspace}
\newcommand{\featureLoss}{\ensuremath{\mathcal{L}_{f}}\xspace}
\newcommand{\styLoss}{\ensuremath{\mathcal{L}_{s}}\xspace}
\newcommand{\imgLoss}{\ensuremath{\mathcal{L}_{img}}\xspace}
\newcommand{\warpLoss}{\ensuremath{\mathcal{L}_{w}}\xspace}
\newcommand{\Loss}{\ensuremath{\mathcal{L}}\xspace}

\newcommand{\pixelW}{\ensuremath{\mathcal{\lambda}_{p}}\xspace}
\newcommand{\featureW}{\ensuremath{\mathcal{\lambda}_{f}}\xspace}
\newcommand{\styW}{\ensuremath{\mathcal{\lambda}_{s}}\xspace}
\newcommand{\warpW}{\ensuremath{\mathcal{\lambda}_{w}}\xspace}

\newcommand{\imgSize}{\ensuremath{\textit{s}_{\textit{I}}}\xspace}

\newcommand{\feature}{\ensuremath{\textbf{\textit{f}}_{j}}\xspace}
\newcommand{\featureM}{\ensuremath{\textbf{\textit{m}}_{j}}\xspace}
\newcommand{\gram}{\ensuremath{\textbf{\textit{g}}_{j}}\xspace}

\newcommand{\featureSize}{\ensuremath{\textit{s}_{f_{j}}}\xspace}
\newcommand{\featureChannel}{\ensuremath{\textit{c}_{j}}\xspace}

\newcommand{\adptGazeMethod}{\textit{\textbf{U-LinFT}}\xspace}
\newcommand{\linearGazeMethod}{\textit{\textbf{U-Lin}}\xspace}
\newcommand{\svrGazeMethod}{\textit{\textbf{U-SVR}}\xspace}
\newcommand{\ssvrGazeMethod}{\textit{\textbf{S-SVR}}\xspace}
\newcommand{\directGazeMethod}{\textit{\textbf{DTrain}}\xspace}

\newcommand{\directGazeMethodAll}{\textit{\textbf{DTrain} (ResNet, full data)}\xspace}

\newcommand{\UnsupAllHP}{\textit{\textbf{U-Train}}\xspace}
\newcommand{\directGazeMethodAllHP}{\textit{\textbf{DTrain}}\xspace}

\newcommand{\GResNet}{\textbf{\textit{ResNet}}\xspace}
\newcommand{\VGG}{\textbf{\textit{VGG}}\xspace}
\newcommand{\VGGSix}{\textbf{\textit{VGG16}}\xspace}
\newcommand{\MnistNet}{\textbf{\textit{MnistNet}}\xspace}
\newcommand{\GResNetHP}{\textbf{\textit{ResNet+HP}}\xspace}

\newcommand{\HCS}{\textbf{\textit{HCS}}\xspace}
\newcommand{\WCS}{\textbf{\textit{WCS}}\xspace}

\newcommand{\mypartitle}[1]{\vspace*{0.75mm}{\noindent {\bf #1}}}

\cvprfinalcopy 

\def\cvprPaperID{9594} 

\ifcvprfinal\fi
\begin{document}

\title{Unsupervised Representation Learning for Gaze Estimation\\[-4.5mm]}

\author{
  Yu Yu,  \ \ Jean-Marc Odobez\\[-0.3mm]
Idiap Research Institute, CH-1920, Martigny, Switzerland\\[-0.3mm]
EPFL, CH-1015, Lausanne, Switzerland \\ [-0.3mm]
 \{yyu,  odobez\}@idiap.ch\\[-2mm]
}


\maketitle

\begin{abstract}
\vspace{-1.2em}
Although automatic gaze estimation is very important to a large variety of application areas,
it is difficult to train accurate and robust gaze models,
in great part due to the difficulty in collecting large and diverse data (annotating 3D gaze is expensive
and existing datasets use different setups).
To address this issue, our main contribution in this paper is to  propose  an effective
approach to learn a low dimensional gaze representation without gaze annotations,
which to the best of our best knowledge, is the first work to do so. 
%
The main idea is to rely on a gaze redirection network and use the gaze representation difference of the input  and  target images (of the redirection network) as the redirection variable.
A redirection loss in image domain allows the joint training of both the
redirection network and the gaze representation network.
In addition, we propose a warping field regularization  which not only
provides an explicit physical meaning to the gaze representations but also avoids redirection distortions.
Promising results on few-shot gaze estimation
(competitive results can be achieved with as few  as $\leq100$ calibration samples),
cross-dataset gaze estimation, gaze network pretraining, and another task (head pose estimation)
demonstrate the validity of our framework.

\vspace{-1em}

\end{abstract}

\section{Introduction}
\vspace*{-2mm} 

Gaze is a non-verbal cue with many functions.
It can indicate attention, intentions, serve as communication cue in interactions,
or even reveal higher level social contructs of people in relation with their personality.
As such, it finds applications in many areas.
For instance, it can be used in multi-party interaction analysis \cite{Ba:ICASSP:2008}, Human-Robot-Interaction (HRI) for both
floor control analysis and for robot behaviour synthesis to enable smooth interactions~\cite{Andrist:2014:CGA:2559636.2559666,Sheikhi:PRL:2015};
in the virtual reality  industry~\cite{konrad2019gazecontingent, chen2019study},
visual rendering can be improved by infering the user gaze direction;
in psychology, gaze behavior can contribute to  mental health  analysis and care~\cite{vidal12_comcom, huang2016stressclick}.

\begin{figure}[tb]
  \centering
  \includegraphics[width=70mm]{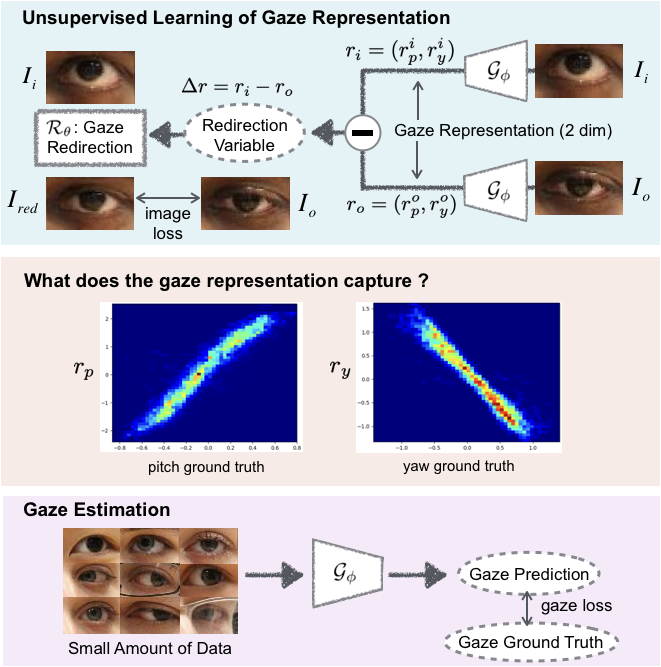}
  \caption{\small{Proposed framework.
    { \it Top:} the networks \GNet extracts gaze representations from two input eye images \inEye and \outEye. 
    Their  difference \deltaRep  is used as input to a gaze redirection network \RNet along with the input image \inEye
     to generate a redirected eye \redEye which should be close to \outEye. Both the \GNet and \RNet networks
      are trained jointly in an unsupervised fashion  from unlabbeled image pairs $(\inEye, \outEye)$.
      {\it Middle.} Thanks to our warping field regularization,
      the distribution of ($\repP$ vs pitch) and ($\repY$ vs yaw) exhibit high (almost linear) correlation.
      {\it Bottom.} The network \GNet can further be used to train a gaze regressor.}
  }
  \vspace{-1em}
  \label{fig:introduction_framework}
\end{figure}

As with other computer vision tasks, the developments in deep neural networks have
largely contributed to the  progresses made on gaze
estimation~\cite{Zhang2015, Zhu_2017_ICCV,  Cheng_2018_ECCV, Yu_CVPR_2019, park2019fewshot, chen2019appearancebased_2019}.
For all these approaches, it is common sense that their performance depends to a large extent on the available amount of data.
Unfortunately, collecting and annotating  3D gaze data is complex and expensive, which introduces 
%
challenges and problems for gaze estimation, as summarized below:
%
\begin{compactitem}
\item \textbf{Data amount.}
  %
  The size of benchmark datasets~\cite{Zhang2015, zhang2015appearance, funes2014eyediap, smith2013gaze, Fischer2018, sugano2014learning}
  including the  number of people, is limited,   
%
%
making it difficult to train robust person-independent models.
%
Synthetic  data~\cite{Wood2015, wood2016learning} offers an alternative, but the domain gap is hard to eliminate.
%
%
\item \textbf{Data annotation.} 3D gaze annotation can be noisy, due to
  (i) measurement errors: most datasets compute the 3D line of sight
  by visually estimating the 3D positions of eyes and gaze targets;
  (ii) participant  distractions  or blinks~\cite{Siegfried:2017:TUS:3136755.3136793, CortaceroICCV2019W},
  leading to  wrong annotations.
\item \textbf{Dataset bias.}
  Existing datasets rely on different cameras and setups, with important variations in  visual appearances
  (resolutions, lighting conditions).
  More importantly, they may only provide eye images obtained using different preprocessing techniques and gaze coordinate systems,
  making it almost impossible to merge datasets for training.
  It is thus hard to apply trained model to out-of-domain samples.
\end{compactitem}
To address these challenges and lower the requirements for annotated gaze dataset,
we propose an unsupervised approach which leverages large amounts of 
\textbf{unannotated eye images} for learning  gaze representations,
and  only a \textbf{few calibration samples} to train a final gaze estimator.
We show in experiments that with as low as 100 calibration samples
we  can already achieve competitive performances.

The main idea is illustrated in Fig.~\ref{fig:introduction_framework}.
The basis is a redirection network \RNet which takes as input an eye image \inEye as well as a
gaze redirection variable \deltaV.
It generates an output image \redEye of the same eye but with the redirected gaze. 
In prior works~\cite{Ganin2016, Kononenko2017, Yu_CVPR_2019, He_2019_ICCV_copy}, \deltaV is explicitly set as a gaze offset,
which means that gaze annotated images are required at training time
(to set the gaze difference between \inEye and the target output image \outEye).
In contrast, our method aims at using a network \GNet to extract gaze representations
from \inEye and \outEye and the simple representation difference provides the sufficient information required to do gaze retargeting. 
%
%
By imposing appropriate loss functions between the redirected output $\redEye$ and the target $\outEye$,
the framework can jointly train both the \RNet and \GNet networks from unlabelled images,
implicitly enforcing the unsupervised learning of gaze representations.
The middle part of Fig.~\ref{fig:introduction_framework} shows that this is achieved, as the  2-dimensional output of \GNet
is highly correlated (close to linear correlation) with groundtruth gaze angles.
It is then possible to train a robust gaze estimator leveraging this representation.
While investigating the above ideas, this paper makes the following contributions:
\begin{compactitem}
\item \textbf{Unsupervised gaze representation learning.} 
  We propose an approach to learn low dimensional gaze representations without gaze annotations,
  relying on a gaze redirection network and loss functions in image domain.
  To our best knowledge, this is the first work of unsupervised gaze representation learning. 
  %
  %
\item \textbf{Warping field regularization.}
  Similar to previous works, we rely on an inverse warping field \Warp to perform gaze redirection. 
  This paper proposed a warping field regularization which not only prevents possible overfitting or distortions,
  but also gives a physical meaning to the components of the learned unsupervised gaze representations.
\item \textbf{Head pose extensions.}
  We also show that our unsupervised method is not limited to gaze estimation,
  but can also be used to process face images and learn a head pose related representation.
\end{compactitem}
Experiments on three public datasets demonstrate the validatity of our approach, 
in particular when training with very few gaze calibrated datapoints
and applying to cross-domain experiment (which shows that our method could successfully leverage large amount of Internet data to handle a much larger variety of eye shape, appearance, head poses, and illumination, ending in a more robust network for gaze representation extraction).

%

In the rest of the paper,
we first summarize related works in Section 2.
The method is detailed in Section 3.
Section 4 explains our experiment protocol and reports
our results. The conclusion is drawn in Section 5.

\section{Related Work}
\vspace*{-2mm} 

Gaze estimation can be categorized into 3 classes, 2D Gaze Estimation, Gaze Following and 3D Gaze Estimation. 


\mypartitle{2D Gaze Estimation} 
aims at predicting the 2D fixation point of gaze, 
%
e.g. on the screens of mobile devices~\cite{Krafka2016, huang2015tabletgaze}. 
They usually rely on large datasets since annotating 2D gaze data is efficient. 
But it is hard to generalize a 2D gaze model to multiple devices or scenarios.


\mypartitle{Gaze Following }
attempts to infer the object people are looking at.  
Recasens et al. proposed to use a saliency  and a gaze pathways to predict the objects people look at
in an image~\cite{Recasens2015} or in a video~\cite{8237422}.
Gaze following models tend to predict the head pose rather than the gaze,
although recent works~\cite{Chong_2018_ECCV} attempted to jointly model gaze following and
3D gaze estimation, but without much improvement.

\mypartitle{3D Gaze Estimation}
which retrieves the 3D line of sight of eyes is the main focus of this paper.
Traditional approaches mainly include geometric based methods (GBM) and appearance based methods (ABM). 
GBM methods first extract features~\cite{venkateswarlu2003eye,FunesMora2014,Wood2016,ishikawa2004passive, Wood2014, Gou_eye, Gou2017, TiBa11b, Villanueva2013} from training images then estimate parameters of a geometric eye model which could be used to predict gaze.
They usually require high resolution eye images and near frontal head poses, which limits its application scope. 
In contrast, ABM methods~\cite{hansen2010eye, tan2002appearance, huang2015tabletgaze, noris2011wearable, martinez2012gaze, sugano2014learning, Lu2011a, FunesMora2016} learn a direct mapping from eye appearance to the corresponding gaze.

ABM methods have attracted more attention in recent years with the development of deep learning.
Zhang et al.~\cite{Zhang2015} proposed a shallow network combining head pose
along with extracted eye features, and later showed that a deeper network
can further improve  performance~\cite{Zhang2017a}. 
Moving beyond single eye gaze estimation,
Cheng et al.~\cite{Cheng_2018_ECCV} proposed to use two eyes  while others
relied on the full face, like in  Zhang et al.~\cite{zhang2015appearance}
where a network process a full face
but without using the head pose  explicitly.
%
Zhu et al.~\cite{Zhu_2017_ICCV}, however, proposed a geometric transformation layer to model the gaze and head pose jointly. 
Finally, as a recent trend, researchers start to work on building person-specifc models from few reference samples 
to eliminate the person specific bias~\cite{Liu2018, DBLP_journals_corr_abs-1904-09459, Yu_CVPR_2019, park2019fewshot,
  chen2019appearancebased_2019,  Xiong_2019_CVPR, lindn2018learning}.
%

In general however, the performance of all the above models depends on the amount and diversity of training data.
But as annotating 3D gaze is complex and expensive, it is difficult to collect data. 
Although synthetic data and domain adaptation~\cite{wood2016learning, Wood2015, journals_corr_ShrivastavaPTSW16, Wang2018, yudeep} have been proposed, the domain gap between the synthetic data and real data is difficult to eliminate.

\mypartitle{Representation Learning } is also a topic related to our paper.
%
Wiles et al.~\cite{DBLP_conf_bmvc_KoepkeWZ18} proposed FAb-Net which learns a face embedding
by retargetting the source face to a target face. The learned embedding  encodes facial attributes
like head pose and facial expression.
Li et al.~\cite{Li_2019_CVPR} later extended this work by disentangling the facial expression
and the head motion through a TwinCycle Autoencoder. The training of the two approaches are conducted in a self-supervised way. 
Different from the above approaches which  learn high dimensional embeddings with unclear physical meaning,
our framework learns a low dimensional representations (2-Dim) with very clear meaning. 
Finally, following the face retargetting framework, an interesting gaze representation learning approach
is proposed by Park et al.~\cite{park2019fewshot} where a face representation extracted
from a bottleneck layer is disentangled as three components:
appearance, gaze and head pose.
The method however used a supervised approach for training,
relying on  head pose and gaze labels.

\section{Method}
\vspace*{-2mm} 

\subsection{Method Overview}
\vspace*{-2mm} 


\begin{figure}[tb]
  \centering
  \includegraphics[width=70mm]{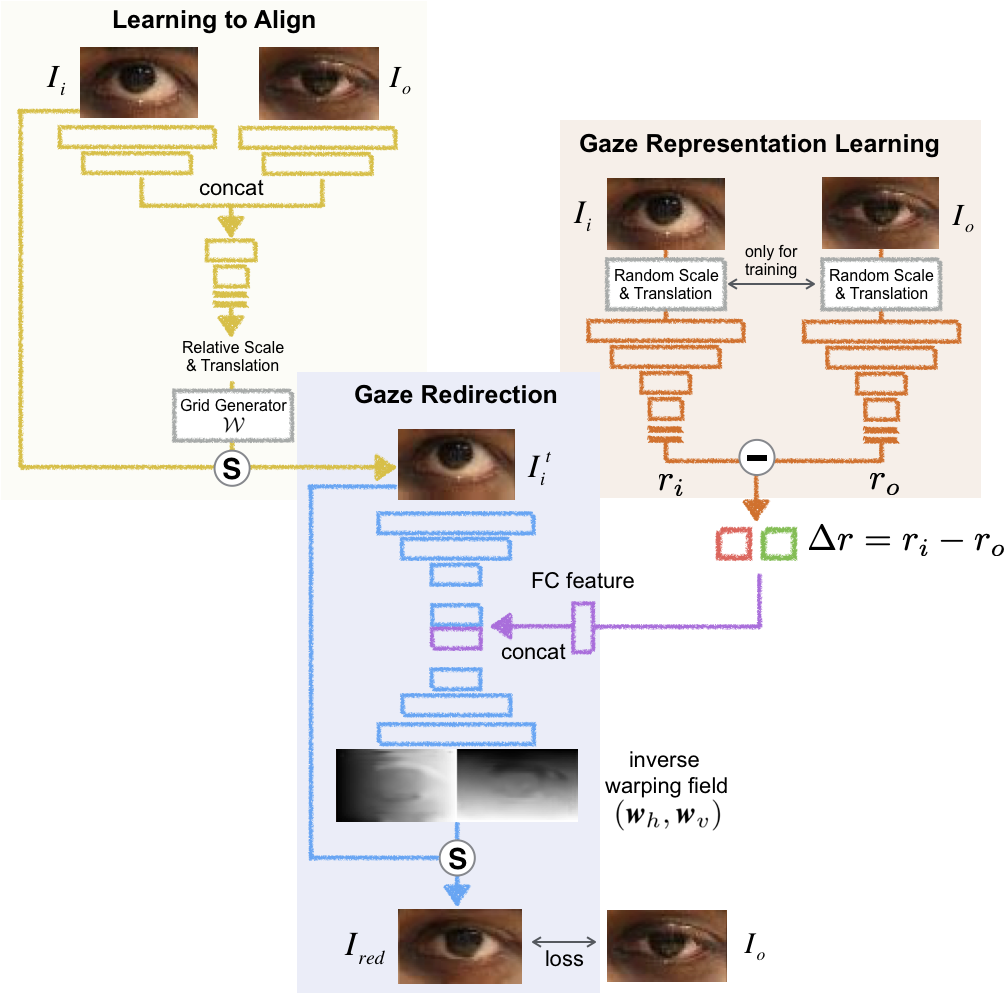}
  \caption{\small{Unsupervised Learning of Gaze Representation.}}
  \vspace{-1em}
  \label{fig:unsup_gaze_est}
\end{figure}

The main idea behind our approach was introduced  in Fig.~\ref{fig:introduction_framework}:
the aim is to  jointly learn a representation network \GNet and a redirection network \RNet so that
the difference $\deltaRep = \rep_i - \rep_o = \GNet(\inEye) -  \GNet(\outEye)$ between the extracted gaze representations
indicates the gaze change to be used by the redirection network to generate a redirection image \redEye which is as close as possible to \outEye.

Our more detailed framework is shown in Fig.~\ref{fig:unsup_gaze_est}.
Given a training image pair $(\inEye, \outEye)$ the network does the following.
An alignment network \LNet  aligns the input \inEye to \outEye
using a global parametric motion model (translation and scale) according to:
$\transEye = \LNet(\inEye,\outEye)$.
Then the redirection network  takes this image as input, and produces a retargeted image as
$\redEye = \RNet(\inEye^t,\deltaRep)$, where as above \deltaRep denotes the intended gaze change
to be applied.
In the following, we further motivate and detail our three networks.
Then, we introduce the loss used for training the system, with
a particular attention paid to the regularization of the warping field involved in the gaze redirection.
Note that as a requirement for gaze redirection~\cite{Yu_CVPR_2019},
the image pair $(\inEye, \outEye)$ should be from the same person and share a similar head pose.



\vspace*{-1mm} 
\subsection{Gaze Representation Learning}
\label{sec:gaze_representation}
\vspace*{-2mm} 

The top right  of Fig.~\ref{fig:unsup_gaze_est} shows the architecture of the gaze representation learning module.
It first extracts the gaze representations from the input images with $\GNet$,
a network based on ResNet blocks, and then computes the representation difference.
In our approach, there are several elements which favor the learning of a gaze related representation rather than other information.

\mypartitle{Gaze Representation \rep.} 
We set \rep to be of dimension 2, which is motivated by two aspects.
First, as the gaze direction is defined by the pitch and yaw angles, a 2D representation is enough.
Secondly, a  compact representation avoid the risk of capturing appearance
information which should be extracted by the encoding part of the redirection network \RNet from the input image.
Otherwise,  with a higher dimension, both \RNet and $\GNet$ may encode eye appearance features,
making the training of \RNet and $\GNet$ less constrained.
%

\mypartitle{Data Augmentation.}
To further enforce \GNet to capture gaze-only information, we assume that the gaze representation should remain the same under small geometric pertubations.
Thus, during training, we also apply random scaling and translation to the images before applying \GNet.
This data augmentation is a key step to achieve robust and accurate unsupervised gaze learning.
Should  this step be removed,  $\GNet$ might learn to detect the pupil center position,
which would be sufficient for  the network $\RNet$ to generate a precise redirection output,
but not be what we want.
Thus,  data augmentation enforces $\GNet$ to learn a scale and translation invariant representation, i.e. a gaze representation.


\vspace*{-1mm} 
\subsection{Global Alignment Network \LNet}
\vspace*{-2mm} 

As pointed in~\cite{Yu_CVPR_2019}, training a gaze redirection network requires well aligned eye image pairs
since global geometric transformation information can not be retrieved from the input image or the gaze difference.
Hence, previous works used synthetic data~\cite{Yu_CVPR_2019} (domain adaptation required), landmark detection~\cite{Zhang2015} or 3D head model~\cite{Funes-Mora2012} for eye alignment, which is not precise enough. 
Inspired by~\cite{NIPS2015_5854}, we propose to learn to align an input image $\inEye$
with a target output  $\outEye$, as shown in the top left  of Fig.~\ref{fig:unsup_gaze_est}.
Concretely, an alignment sub-network $\LSubNet$ takes $\inEye$ and $\outEye$ as input and predicts
the motion parameters (translation and relative scale) between $\inEye$ and $\outEye$.
In the first few layers of \LSubNet, the two images are processed by separate network branches with shared weights.
Then the extracted  image features are concatenated and further processed  to predict  the geometric parameters.
A grid generator~$\gGen$ ~\cite{NIPS2015_5854} then converts these parameters
into the inverse warping field  transforming $\inEye$ into  $\transEye$ (supposed to be aligned  with $\outEye$).
The whole forward process can be formulated as: \\[-2.5mm]
\begin{equation}
\transEye = \LNet(\inEye, \outEye ) = \inEye \circ \gGen ( \LSubNet (\inEye, \outEye ))
\label{eq:eye_trans}
\end{equation}
\vspace*{-3mm}

\noindent where $\circ$ denotes the grid sampling operator.  Fig.~\ref{fig:unsup_gaze_est}
illustrate one alignment example, where  $\inEye$ has been translated vertically to align with $\outEye$.

\vspace*{-1mm} 
\subsection{Gaze Redirection Network \RNet}
\vspace*{-2mm} 

The network \RNet is shown in the bottom part of Fig.~\ref{fig:unsup_gaze_est}.
The main part is an encoder-decoder network \RSubNet trained to predict a warping field $\Warp =(\xWarp, \yWarp)$
which  warps  the (aligned) input  $\transEye$ using a grid sampling operation~\cite{NIPS2015_5854}
and synthesize a gaze redirection output $\redEye$.
%
In its bottleneck part, the network also receives feature maps
generated from the retargeting gaze information \deltaRep between \inEye and \outEye.
As discussed in Sec.~\ref{sec:gaze_representation}, the encoder of \RSubNet ought to encode the eye structure (appearance) related
information of \transEye, while \GNet (through \deltaRep) should encode only the gaze change.
The whole forward process can be summarized as:\\[-2.5mm]
%
\begin{equation}
\redEye=\RNet(\transEye,\deltaRep) = \transEye \circ \RSubNet(\transEye, \GNet(\inEye) - \GNet(\outEye))
\label{eq:eye_trans_red_rep}
\end{equation}

\vspace*{-1mm} 

\subsection{Training Loss, Warping Field Regularization}
\vspace*{-1mm} 

The loss used to  train the whole system is defined as a linear combination of several losses:\\[-2.5mm]
\begin{equation}
  \Loss = \imgLoss + \warpW \warpLoss \mbox{ with } \imgLoss = \pixelW \pixelLoss + \featureW \featureLoss + \styW \styLoss
  \label{eq:loss}
\end{equation}
where  \imgLoss is an image loss defined at the pixel (\pixelLoss), feature (\featureLoss), and style (\styLoss) levels,
whereas \warpLoss is a regularization term on the warping field.
In the following, we first introduce \imgLoss, and then
emphasize the warping loss \warpLoss which plays an important role in our approach.

\vspace*{-2mm} 
\subsubsection{Image Loss \imgLoss}
\vspace*{-2mm} 

The main goal of the image loss is to measure the semantic difference between the generated image $\redEye$
and the target image  $\outEye$. It comprises three terms that we now describe.

\mypartitle{Pixel Loss.}
It measures the discrepancy between $\redEye$ and $\outEye$ using a pixel level L1 loss ($\imgSize$ denotes the image size).

\vspace{-3mm}
\begin{equation}
\pixelLoss = \frac{1}{\imgSize}||\redEye - \outEye||_{1}
\label{eq:pixel_loss}
\end{equation}
\vspace{-2mm} 

\mypartitle{Perceptual Loss.}
\pixelLoss is local and sensitive to illumination differences, and 
does not capture more structure and semantic information disparities.
%
The latter and the robustness to illumination changes
can be achieved using a perceptual loss comprising both feature and style reconstruction losses~\cite{Johnson2016Perceptual}
%
which can be computed as follows.
The $\redEye$ and $\outEye$ images are passed through a VGG16 network  pretrained with ImageNet,
from which we consider the features $\textbf{\textit{f}}_j$ in the  $j=3, 8,$ and $13^{th}$  layers.
Accordingly, we can define the feature reconstruction loss \featureLoss as:\\[-2.5mm]
\begin{equation}
  \featureLoss = \sum_{j} \frac{1}{\featureChannel \cdot \featureSize}||\feature(\redEye) - \feature(\outEye)||_{2}
\label{eq:feature_loss}
\end{equation}
\vspace{-4mm}

\noindent in which  $s$ represents the spatial size of the feature maps and $c$ the number of feature channels.
To compute the style loss \styLoss, the 3D feature maps $\feature$ are first reshaped into
2D matrices $\featureM$ of size $\featureChannel \times \featureSize$
from which we can compute the gram matrices $\gram$ (size $\featureChannel \times \featureChannel$), and then \styLoss: \\[-2.5mm]
\begin{equation}
  \mbox{\hspace{-2mm}}
  \gram \! = \! \frac{1}{\featureSize} \featureM \! \cdot \! \featureM^{T}, \, 
  \mbox{and }
  \styLoss \! = \! \sum_{j} \! \frac{1}{\featureChannel^{2}} ||\gram(\redEye) - \gram(\outEye)||_{2}  . \! \! 
  \label{eq:styloss}
\end{equation}
\vspace{-3mm}




\vspace*{-2mm} 
\subsubsection{Warping Field Regularization}
\vspace*{-2mm} 

\begin{figure}[tb]
  \centering
  \includegraphics[width=80mm]{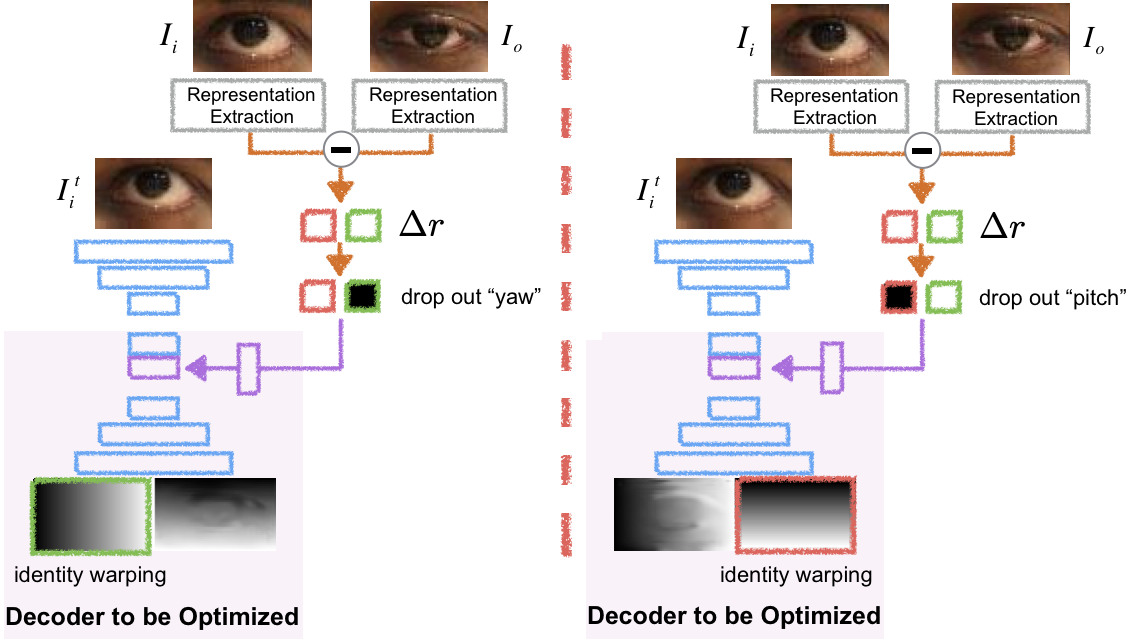}
  \caption{\small{Representation drop out for warping field regularization.}}
  \vspace{-1em}
  \label{fig:warp_reg}
\end{figure}


\begin{figure*}[tb]
  \hspace{-1.1em}\subfigure{
  \scriptsize{a)}\hspace{-1.3em}\includegraphics[height=22mm]{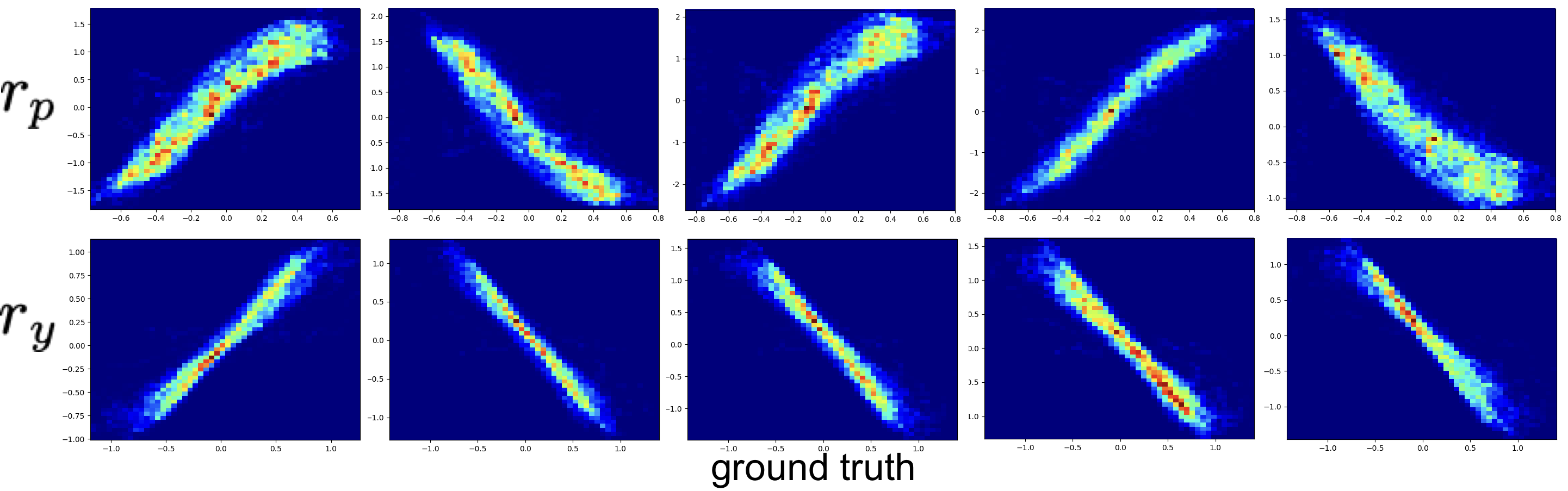}
  }
  \hspace{-0.7em}\subfigure{
  \scriptsize{b)}\hspace{-0.7em}\includegraphics[height=22mm]{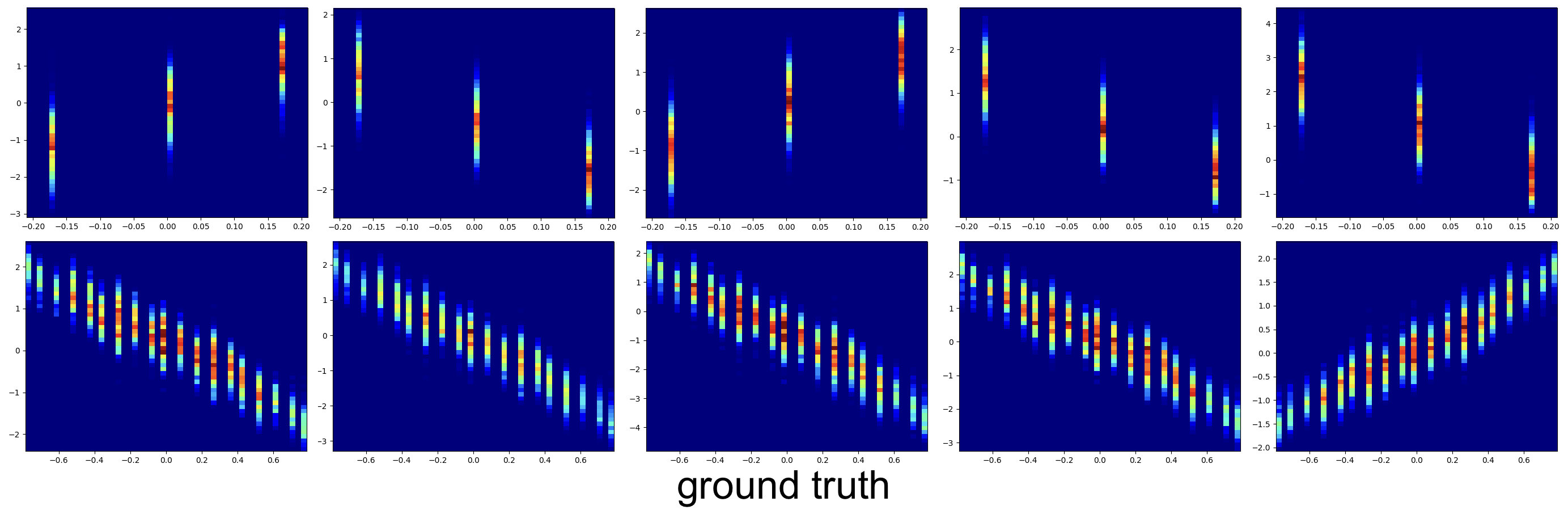}
  }
  \hspace{-0.7em}\subfigure{
  \scriptsize{c)}\hspace{-0.7em}\includegraphics[height=22mm]{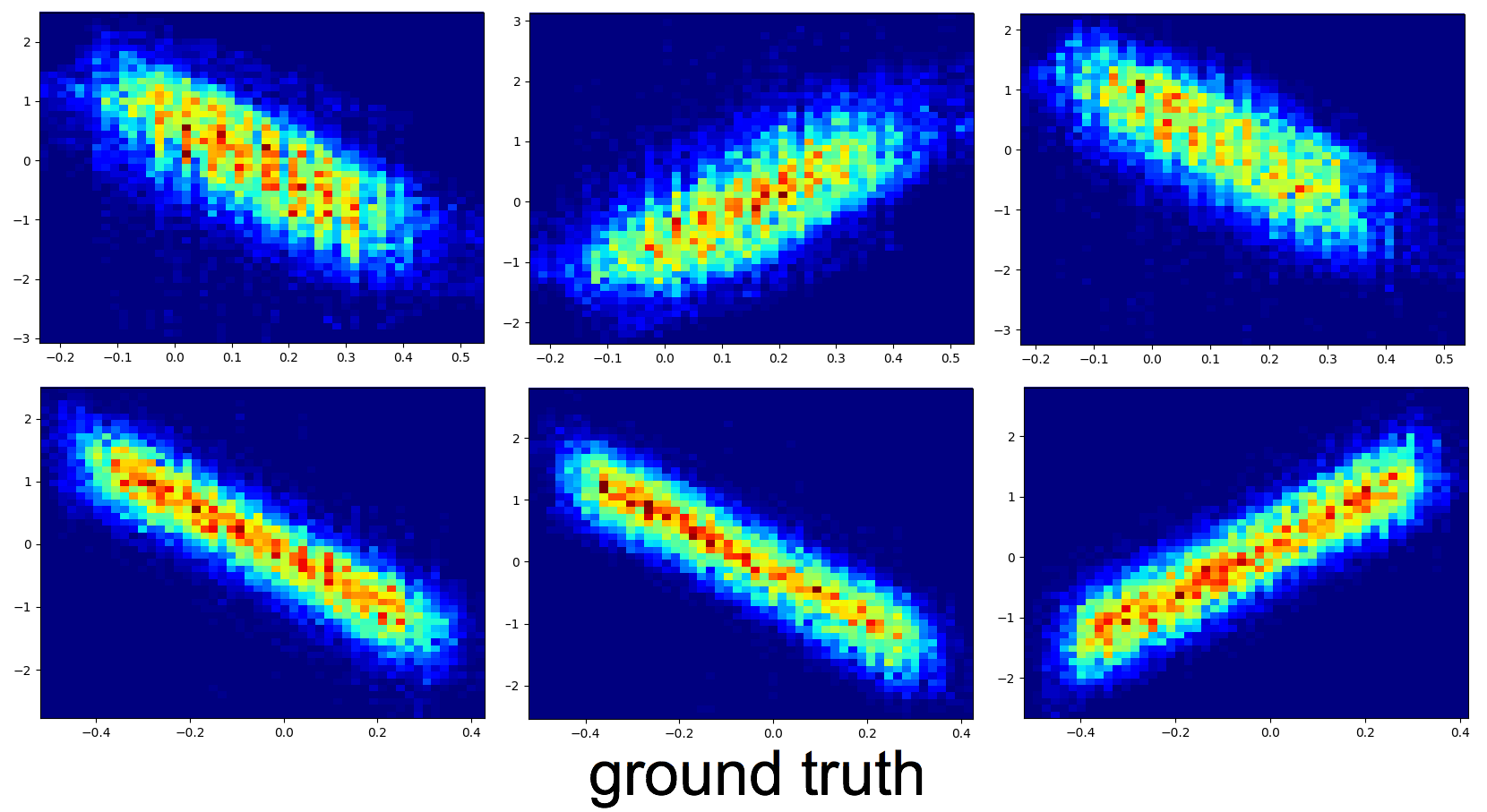}
  }
  \vspace{-1em}\caption{\small{Distribution of the $(pitch, \repP)$ and $(yaw, \repY)$ data points (with $(yaw, pitch)$ being
    the  ground truth gaze in head coordinate system ($\HCS$)), for different training folds of a dataset
    (a) Eyediap. (b) Columbia Gaze (discrete and sparse gaze label). (c) UTMultiview.}
  }
  \vspace{-1.5em}
  \label{fig:unsup_vis_rsl}
\end{figure*}

\mypartitle{Motivation.}
With the image loss $\imgLoss$ alone, we are able to train the whole framework in an end-to-end fashion
and achieve the unsupervised learning of gaze representation.
However, the physical meaning of the gaze representation is not clear.
In this section, we introduce a warping field regularization
which not only gives physical meaning to the gaze representation
but also regularizes the training of the whole framework (as shown in the experiments).

The main idea is to associate each gaze representation component  with a specific warping field.
Indeed, as shown in ~\cite{yudeep}, the gaze yaw mainly involves an horizontal motion of the iris,
and the pitch  a vertical motion of the eyelid and iris.
In other words, when there is only a yaw change (no pitch change), the vertical motion flow of eye region
should be close to 0. Similarly, with only a pitch change, the horizontal flow should  be close to 0.
Note that no motion flow corresponds to an identity  warping field.

\mypartitle{Gaze Representation Dropout.}
To exploit the above assumption, we proposed the dropout mechanism illustrated in  Fig.~\ref{fig:warp_reg},
in which we drop in turn each dimension of the gaze change $\deltaV$ (setting it to 0),
and enforce one of the warping field components to be an identity mapping $\idWarp$,
while keeping the other one unchanged.

More concretely, given a training image pair $(\inEye,\outEye)$, we first apply the forward pass and compute
the representation difference as well as the warping field, according to:
\begin{equation}
  [\deltaGazeP, \deltaGazeY]=\GNetInfer(\inEye) - \GNetInfer(\outEye), 
(\xWarp, \yWarp)=\RSubNetInfer(\transEye, [\deltaGazeP, \deltaGazeY])
\label{eq:eye_rep_infer}
\end{equation}
Then, we apply the dropout for each dimension, which results in the fields:
\begin{equation}
\begin{aligned}
&(\xWarp^{\deltaGazeY=0}, \yWarp^{\deltaGazeY=0})=\RSubNetDecode(\transEye, [\deltaGazeP, 0])\\
&(\xWarp^{\deltaGazeP=0}, \yWarp^{\deltaGazeP=0})=\RSubNetDecode(\transEye, [0, \deltaGazeY])
\label{eq:eye_do_rep_infer}
\end{aligned}
\end{equation}
%
%
on which we apply the regularization loss:
\begin{equation}
\hspace{-1em}
\begin{aligned}
\warpLoss = \frac{1}{\imgSize}(&||\xWarp^{\deltaGazeY=0} - \idWarp||_{1} + ||\yWarp^{\deltaGazeY=0} - \yWarp||_{1} + \\
&||\yWarp^{\deltaGazeP=0} - \idWarp||_{1} + ||\xWarp^{\deltaGazeP=0} - \xWarp||_{1})
\label{eq:warploss}
\end{aligned}
\end{equation}
Note that for the dropout of each dimension, we not only enforce one field to be identity mapping, but also keep the other field unchanged since the other dimension of $\deltaV$ is unchanged.
In addition, note that for this regularization term, only the parameters $\theta^{*}$ of the decoder
part of the redirection network are  optimized, as shown in Fig.~\ref{fig:warp_reg}.
This  regularization is used along with the image loss when training the network (see Eq.~\ref{eq:loss}).
In essence, through this dropping and regularization mechanism, the network will be trained to associate the generation of
one warping direction with one representation component, giving a physical meaning to the gaze representation.
This is demonstrated in Fig.~\ref{fig:unsup_vis_rsl}.
In addition, as shown in Section~\ref{sec:resqualitative}, this regularization term also
prevents potential overfitting or distortion of the warping field, leading to improved gaze redirection images 
and better gaze representations when used for gaze training.






\subsection{Few-Shot Gaze Estimation}
\vspace*{-2mm} 

\mypartitle{Linear Adaptation.}
To estimate the gaze $(\GazeP, \GazeY)$ (in head coordinate system, $\HCS$) from
the unsupervised gaze representation $(\repP, \repY)$ (also in $\HCS$),
we can first simply estimate two linear models (for pitch and yaw respectively):
\vspace{-0.5em}
\begin{equation}
\GazeP = \kP \repP + \bP  \hspace{0.2em},
\GazeY = \kY \repY + \bY 
\label{eq:linear_gaze_model}
\end{equation}
using the very few calibration samples to rescale our representation, 
where $\kP$, $\bP$, $\kY$ and $\bY$ are model parameters.

\mypartitle{Network Re-initialization and Finetuning.}
The second step is to fine-tune the network using the calibration samples.
However, before doing this we re-initialized the weight and bias
($\kNet = [\kPNet, \kYNet]$, $\bNet = [\bPNet, \bYNet]$) of the last layer of gaze network $\GNet$,
using the above linear models, according to:
\vspace{-0.5em}
\begin{equation}
\begin{aligned}
&\kP (\kPNet \cdot \featureFC + \bPNet) + \bP \rightarrow (\kP \kPNet)\cdot\featureFC + (\kP \bPNet + \bP) \\
&\kY (\kYNet \cdot \featureFC + \bYNet) + \bY \rightarrow (\kY \kYNet)\cdot\featureFC + (\kY \bYNet + \bY) \\[-1.5em]
\label{eq:net_reinit}
\end{aligned}
\end{equation}
where $\featureFC$ is feature forward to the last layer and $[\kP \kPNet, \kY \kYNet]$ and $[\kP \bPNet + \bP, \kY \bYNet + \bY]$ are the new weight and bias.


\mypartitle{Gaze in World Coordinate System ($\WCS$).}
To obtain a final gaze estimation in the  $\WCS$, the estimation  in $\HCS$ is transformed  using the head pose information.

\subsection{Implementation Detail}
\vspace*{-2mm} 



\mypartitle{Hyperparameters and Optimization.} The framework is optimized by Adam with an initial learning rate of $10^{-4}$ and a small batch size of $16$. $10$ epochs are used to train the network and the learning rate is reduced by half every 3 epochs. The default values of loss weights $\pixelW$, $\featureW$, $\styW$ and $\warpW$ are $1.0$, $0.02$, $0.1$, $0.25$ respectively. But for Eyediap samples which are blurry, we set $\pixelW$ to 0.2.

\mypartitle{Activation Function.} To bound the value of gaze representation when training begins, we used $\textit{tanh}$ in the last layer of gaze network $\GNet$. After 2 epochs, we removed the activation function, making the last layer a linear projection.

\section{Experiment}
\vspace*{-2mm} 

\subsection{Experiment Protocol}
\vspace*{-2mm} 

\mypartitle{Dataset.}
We used three public datasets for experiment: Eyediap~\cite{funes2014eyediap}, Columbia Gaze~\cite{smith2013gaze} and UTMultiview~\cite{sugano2014learning}. 
Eyediap  was collected with a RGBD sensor.
It consists of sessions with different illumination conditions, gaze targets and head motion settings.
We selected the session of HD video, condition B, floating target and static head pose for experiment, which results in 5 videos (5 subjects).
Eye images were extracted and rectified to a frontal head pose~\cite{FunesMora2016}. 
Different from Eyediap, the gaze targets in Columbia Gaze and UTMultiview are discrete:
only 7 horizontal and 3 vertical gaze directions in Columbia Gaze,
160 directions in UTMultiview where the 
gaze labels are further smoothed by a reconstruction-and-synthesis strategy. 
Both Columbia Gaze and UTMultiview use camera arrays to take multiview (head pose) samples.
We use all available data for these two datasets (56 subjects and 50 subjects respectively). 

\mypartitle{Cross-Validation.}
For the 3  datasets, we perform $n = 5, 5, 3$-fold cross validation respectively (no subject overlap). 
In each fold, training data is used for  unsupervised learning (without using  gaze annotations)
and then for few-shot gaze estimation by randomly selecting $10$ to $100$ samples with annotations.
Test data is only used for evaluation. 
Please {\bf note that this few shot setting is different from few shot personalization setting}
as in~\cite{Yu_CVPR_2019, park2019fewshot, He_2019_ICCV},
and all  reported results in this paper are cross-subject.

\mypartitle{Training Pair.}
For Columbia Gaze and UTMultiview, the image pairs $(\inEye, \outEye)$ are randomly selected.
However, for Eyediap which covers a larger gaze range, $(\inEye, \outEye)$ are selected temporally by
limiting their time gap within 10$\sim$20 frames.
As already mentioned, $\inEye$ and $\outEye$
should be of the same person with a similar head pose. We used at most 200K pairs for training.

\mypartitle{Network Models.}
Network details are given in the supplementary material.
For the redirection network, it is based on ResNet blocks.
Regarding the gaze network \GNet,
our default architecture $\GResNet$ is based on 4 ResNet blocks.
For comparative experiments, we  tested with $\VGGSix$ pretrained with ImageNet (adopted in~\cite{Zhang2017a})
and with the $\MnistNet$ shallow architecture used in~\cite{Zhang2015}.

%


\mypartitle{Tested Few-Shot Gaze Estimation Methods.}
\begin{compactitem}
\item $\adptGazeMethod$: our approach, consisting of unsupervised representation learning (U), 
   linear adaptation (Lin) and network finetuning (FT), including re-initialization.

 \item $\linearGazeMethod$: the same as above, but without network finetuning.
   A similar linear adaptation strategy was used in~\cite{Liu2018} (but this was for gaze personalization).

 \item $\svrGazeMethod$: unsupervised representation learning followed by SVR adaptation.
   The SVR input features are the concatenation of the gaze representation and the output from the second last layer of $\GNet$.
   A similar approach was used for gaze personalization in~\cite{Krafka2016}.

\item $\directGazeMethod$: a randomly initialized (or ImageNet pretrained for $\VGGSix$) baseline network  directly trained with calibration samples.
\end{compactitem}
Note that since to our best knowledge this is the first work to investigate unsupervised gaze representation learning and cross subject few-shot gaze estimation,
it is difficult to find a state-of-the-art approach for comparison.

\mypartitle{Performance Measure.}
We use the angle (in degree) between the estimated gaze  and the ground truth gaze vectors as error
measure. All reported results are  the average of 10 runs 
(including random selection of calibration samples).

\subsection{Qualitative results}
\label{sec:resqualitative}
\vspace*{-2mm}

\mypartitle{Visualization of the Unsupervised Gaze Representation.}
Fig.~\ref{fig:unsup_vis_rsl} shows the $n$ distributions of unsupervised gaze representation w.r.t. ground truth. Each distribution corresponds to a gaze model
obtained on the left-out folds of the datasets (see \textbf{Cross-Validation} above). 
%
%
%
%
As can be seen, these distributions are almost  linear,
validating  the relevance of our approach and warping regularization loss.
An interesting point is that the gaze representation is inversely proportional to the ground truth sometimes,
which might be due to the random factor during network initialization and  training.

\begin{figure}[tb]
  \hspace{-1em}\subfigure{
  \scriptsize{a)}\includegraphics[height=29mm]{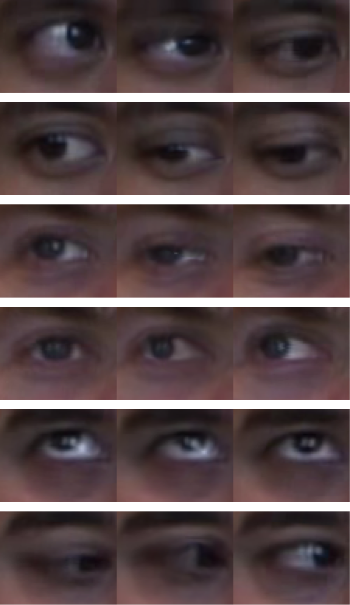}
  }
  \hspace{-0.6em}\subfigure{
  \scriptsize{b)}\includegraphics[height=29mm]{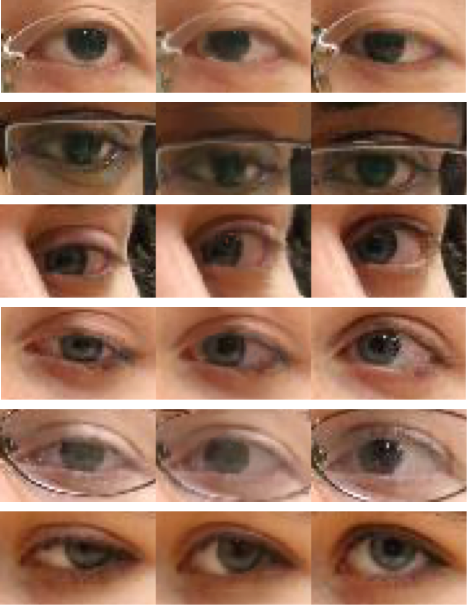}
  }
  \hspace{-0.7em}\subfigure{
  \scriptsize{c)}\includegraphics[height=29mm]{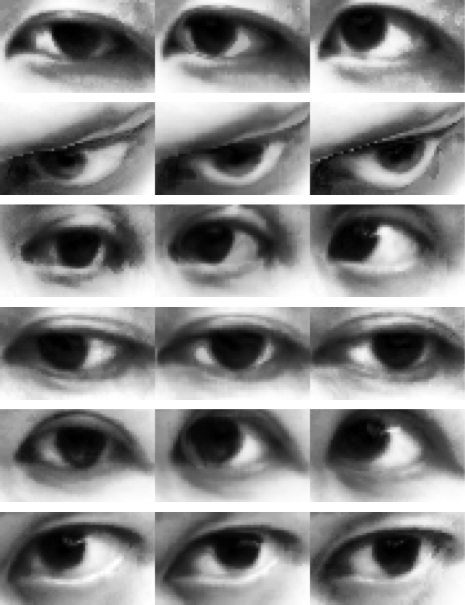}
  }
  \hspace{-0.75em}\subfigure{
  \scriptsize{d)}\includegraphics[height=29mm]{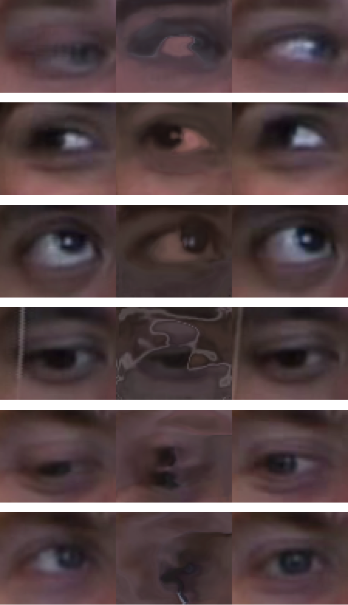}
  }
  \vspace{-1em}\caption{\small{Gaze redirection on
  (a) Eyediap. (b) Columbia Gaze. (c) UTMultiview. (d) Eyediap without the warping regularization loss.
   Each image triple represents: Left: input image, Middle: redirection output, Right: ground truth.}}
  \vspace{-1.5em}
  \label{fig:unsup_gaz_vis_rsl}
\end{figure}

\mypartitle{Gaze redirection.}
Fig.~\ref{fig:unsup_gaz_vis_rsl} illustrates the quality of our unsupervised gaze redirection results,
where we remind that the gaze shift to be applied to the input image
is provided by the representation difference obtained from the left and right images. 
%
As can be seen, our framework achieves accurate gaze redirection as well as eye alignment.
Fig.~\ref{fig:unsup_gaz_vis_rsl}(d) also demonstrates visually the overall benefit of our warping field regularization scheme.

\subsection{Quantitative results}
\vspace*{-2mm} 

\begin{figure*}[tb]
  \centering
  \subfigure{
  \scriptsize{a)}\includegraphics[height=35mm]{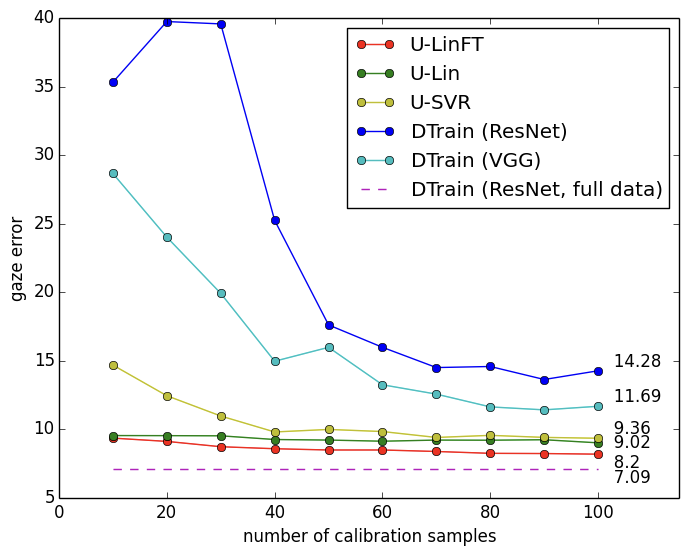}
  }
  \subfigure{
  \scriptsize{b)}\includegraphics[height=35mm]{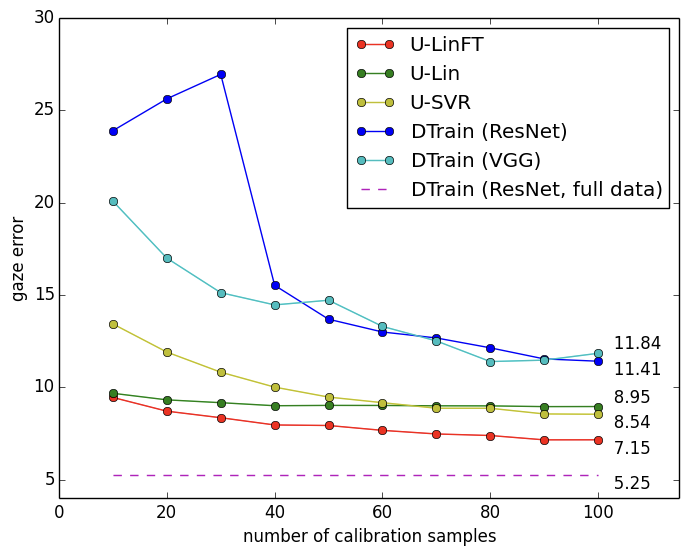}
  }
  \subfigure{
  \scriptsize{c)}\includegraphics[height=35mm]{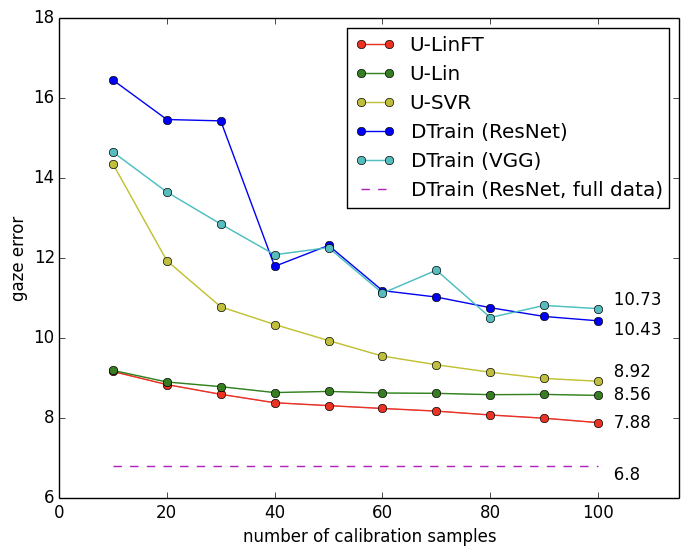}
  }
  \vspace{-0.5em}\caption{\small{Few-shot gaze estimation results. (a) Eyediap. (b) Columbia Gaze. (c) UTMultiview. }}
  \vspace{-1em}
  \label{fig:fs_rsl}
\end{figure*}




\begin{figure*}[tb]
  \begin{minipage}[t]{.42\textwidth}
  \scriptsize{a)
  \hspace{-1.5em}\includegraphics[height=35mm]{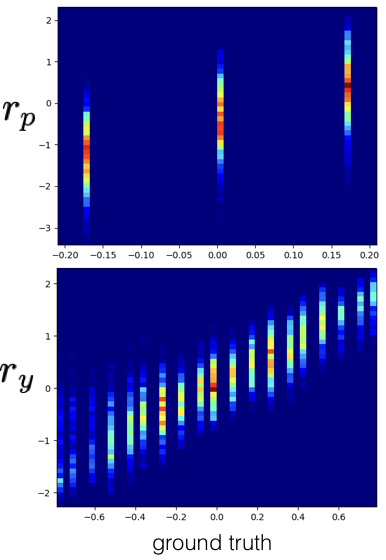}
  }
  \hspace{-1em}\scriptsize{b)
  \includegraphics[height=35mm]{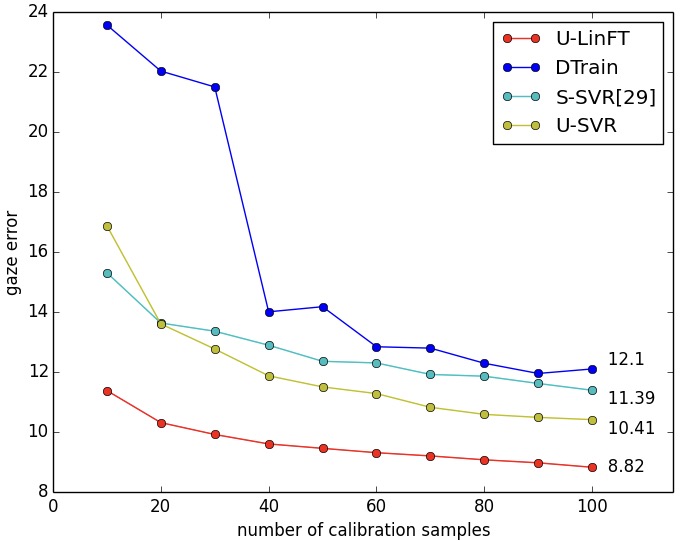}
  }
  \caption{\small{Cross dataset evaluation (unsupervised training on UTMultiview, test on Columbia Gaze)
  (a) Unsupervised gaze representation. (b) Few-shot gaze estimation. }}
  \vspace{-1em}
  \label{fig:cross_fs_rsl}
  \end{minipage}
  \hspace{0.85em}\begin{minipage}[t]{.26\textwidth}
  \includegraphics[height=35mm]{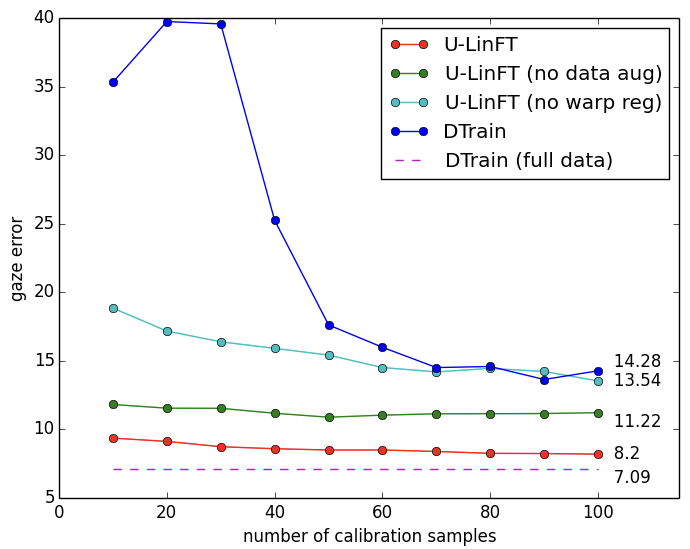}
  \caption{\small{Ablation study (Eyediap). }}
  \vspace{-1em}
  \label{fig:ab_rsl}
  \end{minipage}
  \hspace{0.85em}\begin{minipage}[t]{.26\textwidth}
  \includegraphics[height=35mm]{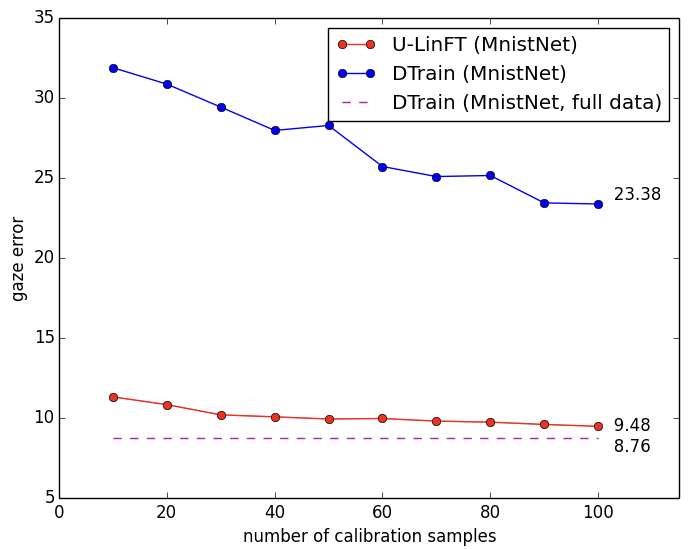}
  \caption{\small{Few-shot gaze estimation with MnistNet. }}
  \vspace{-1em}
  \label{fig:snet_fs_rsl}
  \end{minipage}

\end{figure*}

\mypartitle{Few-Shot Gaze Estimation.}
The quantitative performances of few-shot gaze estimation approaches are reported in Fig.~\ref{fig:fs_rsl},
where the results trained with all data and annotations
($\directGazeMethodAll$) are plot as a lower bound. 
%
We can first notice that our  approach ($\adptGazeMethod$)
achieves an acceptable accuracy ($7^{\circ}\sim8^{\circ}$ error on all datasets) with only 100 calibration samples.
In addition, all few-shot results based on our unsupervised learning ($\adptGazeMethod$, $\linearGazeMethod$, $\svrGazeMethod$)
are significantly better than the $\directGazeMethod$ methods, including the $\VGG$ architecture  pretrained on ImageNet. 
Furthermore, the performance of our  approach with 10 calibration samples is still much better than the performance
of $\directGazeMethod$ methods with 100 samples.

$\adptGazeMethod$ performs very well on Eyediap and UTMultiview
(only about $1^{\circ}$ worse using 100 samples).
%
In contrast, on Columbia Gaze, the  performance gap is $1.9^{\circ}$.
%
One possible reason regarding  Eyediap is that its samples   have been rectified
to a frontal head pose and exhibit less variability,
making our unsupervised gaze learning assumptions more valid.
It is also reflected in Fig.~\ref{fig:unsup_vis_rsl} where the linear distributions from Eyediap are less dispersive. 
This implies that our approach could apply well to  head mounted systems~\cite{DBLP_journals_corr_abs_1905_03702} where eyes
have a fixed pose. 

Amongst the unsupervised approaches,  $\adptGazeMethod$ performs the best.
$\linearGazeMethod$ is in par  using few calibration samples, but the performance gap
increases with the number of samples since linear adaptation has fewer parameters.
%
The $\svrGazeMethod$ method is the worst when using few calibration samples  
because it has to train an SVR model from scratch, but it catches up as the number of  samples increases.





\vspace{-0.5em}
\mypartitle{Cross-Dataset Evaluation.}
We trained our unsupervised model on UTMultiview then tested it on Columbia Gaze (Columbia Gaze samples were converted to grayscale for consistency with UTMultiview).
Fig.~\ref{fig:cross_fs_rsl}(a) visualizes the extracted representations vs ground truth distribution of the Columbia Gaze samples.
%
They still follow a linear-like distribution.
We then randomly select calibration samples from Columbia Gaze for few-shot gaze training.
Results are reported in Fig.~\ref{fig:cross_fs_rsl}(b).
Though we observe a small performance drop compared to results in Fig.~\ref{fig:fs_rsl}(b),
our unsupervised approaches are still much better than the $\directGazeMethod$ method.
%
More interestingly, we also trained a gaze estimator on UTMultiview
in a supervised fashion ($\directGazeMethodAll$) and adapted it on Columbia Gaze. 
This adaptation approach named $\ssvrGazeMethod$ relies on an SVR model which uses features extracted from
the last and the second last layer of the supervised model trained on UTMultiview.
It was used in~\cite{Krafka2016} (the original model only used features
from the second last layer) for cross dataset experiment. 
%
%
Surprisingly, Fig.~\ref{fig:cross_fs_rsl}(b) shows that our unsupervised $\svrGazeMethod$
(based on the same architecture but trained in an unsupervised fashion) is better than $\ssvrGazeMethod$, 
demonstrating that we achieved accurate unsupervised representation learning with good generalization capacity.
These results show that our method can benefit from cross-domain data sources,
and has the capacity to leverage large amount of Internet data to train
robust models coping with  diverse eye shapes, appearance, head poses, and illuminations.


\mypartitle{Ablation Study.}
We study the impact of data augmentation
and warping regularization  by removing them.
%
Note that as the physical meaning of the unsupervised representation is unclear
when removing the warping regularizaton, we used a bilinear model to project the representation to gaze.
Results are shown in Fig.~\ref{fig:ab_rsl}.
%
The performances without data augmentaton or warping regularization are well below our proposed approach,
but  they remain  better than $\directGazeMethod$.
%
The gaze error increases by $\sim \!\! 3^{\circ}$ after removing data augmentation, showing that this is key
to learn scale and translation invariant gaze representations.
But removing the warping regularization leads to even more  performance degradation. 
To explore the cause, some gaze redirection outputs without warping regularization are
shown in Fig.~\ref{fig:unsup_vis_rsl}(d).
In the first three rows, skin pixels are re-projected to the sclera region because of wrong warping.
In the last three rows, the outputs are totally distorted.
This further demonstrates that the warping field regularization
not only gives a physical meaning to the unsupervised gaze representations,
but also prevents from  possible distortions.


\mypartitle{Shallow Architecture.} 
%
As shallow networks can be  of  practical use in mobile devices,
we tested our approach on the  $\MnistNet$ gaze network $\GNet$ while keeping the same $\LNet$ and $\RNet$ architectures.
The performance is shown in Fig.~\ref{fig:snet_fs_rsl}.
Compared with $\GResNet$, the result of $\MnistNet$ is indeed worse.
Nevertheless, we can notice that our approach $\adptGazeMethod$ works much better than the  baseline  $\directGazeMethod$,
and that its performance 
is closer to the lower bound.


\mypartitle{Unsupervised Learning for Pretraining.}
%
In this experiment, we use all the training data and their annotations to fine tune a
model  pretrained in an unsupervised fashion.
%
%
Until now, tested architectures for \GNet (ResNet based or $\MnistNet$)
were taking eye images as input and predicting gaze in $\HCS$
(final gaze obtained by transforming the estimate in $\HCS$  to $\WCS$ with the help of head pose).
Such architectures are suitable for few-shot gaze estimation since the unsupervised gaze representation is also in $\HCS$.  
However, when training with more data and annotations, a better strategy is to predict the gaze in $\WCS$ directly,
by concatenating the convolution features with the head pose before the fully connected layers,
as proposed in~\cite{Zhang2017a}. We denote this architecture as $\GResNetHP$.
Due to this architecture difference in fully connected layers, we only use the convolutional layers of our pretrained $\GResNet$ 
to initialize  $\GResNetHP$,  and randomly initialized the fully connected layers which process the concatenated feature.
Note that since in the Eyediap dataset  eye samples are rectified to frontal head pose, we kept our $\GResNet$ architecture for Eyediap gaze prediction.
%
Tab.~\ref{tab:unsup_pretrain} reports the results.
As can be seen, using our unsupervised training leads to a performance gain of  $0.2^{\circ}\sim0.3^{\circ}$
compared to training from scratch.
This is a small improvement, but given that results are based on 10 rounds of experiments,
it is nevertheless stable and significant. Besides, we also compare our approach with SOTA results in Tab.~\ref{tab:unsup_pretrain}. As can be seen, our approach is better or competitive. Please note that our Eyediap result can not be compared with SOTA directly since we used Eyediap session of HD video, condition B, floating target and static head pose while the listed SOTA used VGA video, condition A, floating target and static head pose. We can not find SOTA results with exactly the same session.
%

\begin{table}[tb]\scriptsize
  \vspace{-5mm}
  \caption{\scriptsize{Gaze estimation with all the training data and annotations using $\GResNetHP$ architecture. $^{\dagger}$p$<$0.01}}
\centering\begin{tabular}{l|lll}
\hline
\diagbox{Model}{Error}{Dataset} & Eyediap & Columbia Gaze & UTMultiview \\ \hline
\hspace{-0.5em}$\UnsupAllHP$ \hspace{-0.5em} & \textbf{6.79}  & \textbf{3.42}   & 5.52        \\ 
\hspace{-0.5em}$\directGazeMethodAllHP$\hspace{-0.5em} & 7.09$^{\dagger}$    & 3.63$^{\dagger}$          & 5.72$^{\dagger}$      \\
\hspace{-0.5em}Yu et al.~\cite{yudeep}  \hspace{-0.5em} & 8.5 & -  & 5.7  \\ 
\hspace{-0.5em}Yu et al.~\cite{Yu_CVPR_2019} (cross subject) \hspace{-0.5em} & - & 3.54  & -   \\ 
\hspace{-0.5em}Zhang et al.~\cite{Zhang2015} \hspace{-0.5em} & -  & -  & 5.9  \\ 
\hspace{-0.5em}Park et al.~\cite{Park2018a} \hspace{-0.5em} & -  & 3.59 & -  \\ 
\hspace{-0.5em}Liu et al.~\cite{DBLP_journals_corr_abs-1904-09459} (cross subject) \hspace{-0.5em} & - & -   & 5.95    \\ 
\hspace{-0.5em}Funes-Mora et al.~\cite{FunesMora2016} \hspace{-0.5em} & 11.6 & -  & -  \\ 
\hspace{-0.5em}Xiong et al.~\cite{Xiong_2019_CVPR} \hspace{-0.5em} & -  & -  & 5.50$\pm$1.03 \\ 
\hspace{-0.5em}Wang et al.~\cite{Wang_2019_CVPR} \hspace{-0.5em} & -  & -  & \textbf{5.4} \\ \hline
\end{tabular}
\label{tab:unsup_pretrain}
\vspace{-2em}
\end{table}

\stepcounter{footnote}\footnotetext{$^{\dagger}$ indicates an error significantly higher than our method ($p<0.01$).}

\begin{figure}[tb]
  \centering
  \includegraphics[height=24mm]{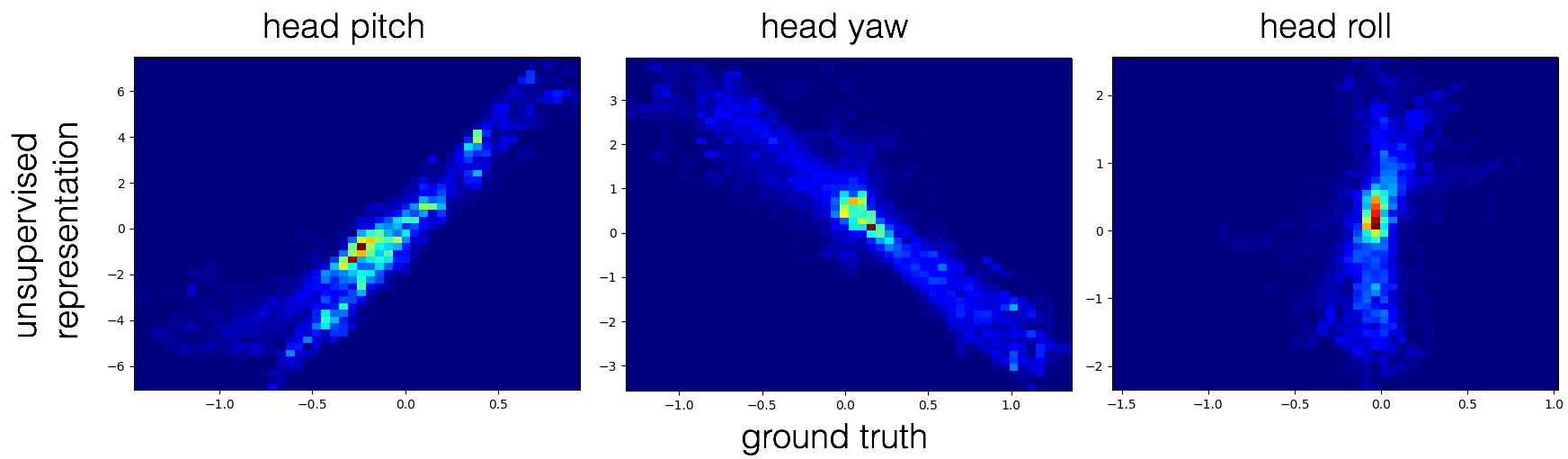}
  \vspace{-0.5em}\caption{\small{Unsupervised head pose representation. }}
  \vspace{-1.5em}
  \label{fig:hp_vis_rsl}
\end{figure}

\mypartitle{Application on Head Pose Estimation.}
We  extended our approach to another task, head pose estimation.
We used cropped faces from BIWI~\cite{fanelli_IJCV} for experiment and selected training pairs randomly within  a temporal window.
As head pose can be described by three rotation angles, pitch, yaw and roll, we used a 3 dimensional  vector to represent it (instead of 2 dim for gaze).
The second change we made concerns the warping field regularization,
where in order to relate  the pitch   with the vertical motion, we enforced the horizontal field to be the identity
when dropping out the representation of yaw and roll; and similarly, when dropping out the
pitch and roll, we defined a loss on the vertical field.
%
%
The unsupervised head pose representation that was learned is illustrated in Fig.~\ref{fig:hp_vis_rsl}.
The distribution of the pitch and yaw representations w.r.t the ground truth still exhibit a high correlation,
but not so much for the roll representation.
There might be two main reasons.
First, none of our regularization terms involves the roll alone;
second, the distribution of rolls in the data is uneven and concentrated, with $80\%$ of them being within $-25^{\circ}\sim5^{\circ}$.
%
Although some future works could be done to improve the unsupervised learning of head pose representation,
we demonstrated the potential of our framework for other tasks.

\section{Conclusion and Discussion}
\vspace*{-2mm}

We have proposed an unsupervised gaze learning framework which,
to the best of our knowledge, is the first work on this topic.
The two main contributing elements are the use of gaze redirection as auxiliary task
for unsupervised learning, and the exploitation of a warping field regularization scheme which not only provides
a  physical meaning to the learned gaze representation dimensions, but also
prevents overfitting or gaze redirection distortions.
We demonstrate promising results on few-shot gaze estimation, network pretraining,
and cross-dataset experiments in which the gaze representation (and network)
learned in an unsupervised fashion proved to be better than a network trained supervisedly with gaze data.
In this view, we believe that our method could successfully leverage internet data to train an unsupervised nework
robust to a large variety of eye shapes, appearance, head poses, and illumination. 

Our work can be expanded along two interesting directions.
First, our work can be used for few shot person-specific gaze estimation.
We believe large performance improvement can be achieved since there is less bias or noise among person specific samples.
It is especially beneficial for few shot gaze estimation (less bias in few data).
Second, given the unsupervised performance on head pose estimation, our work can also be used for full face gaze estimation.
The framework can be implemented by using two network branches, as done in~\cite{Zhu_2017_ICCV}.

Notwithstanding the above advantages, a limitation of our method is that it requires image pairs with the same
or at least close head poses for training, setting some constraints on our approach.
We leave the evaluation of this requirement as a future work.

{\small \mypartitle{Acknowledgement.}
  This work was funded by the European Unions Horizon 2020 research and innovation programme
  -grant agreement no. 688147 (MuMMER, mummer-project.eu).
  }



{\small
\bibliographystyle{ieee_fullname}
\bibliography{egbib}
}

\clearpage

\section{Appendix}

\subsection{Image Resolution}

The image resolution of Columbia Gaze and UTMultiview samples is 36*60,
while the resolution of Eyediap is 60*75.

\subsection{Network Architecture}

\begin{figure}[tb]
  \centering
  \includegraphics[width=50mm]{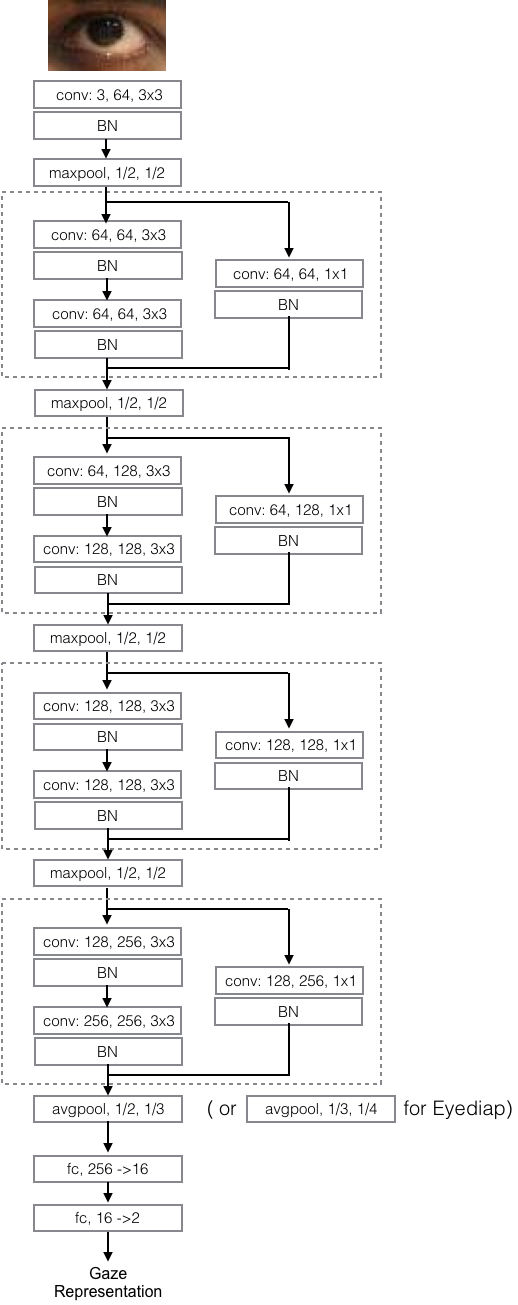}
  \caption{Gaze Representation Learning Network $\GNet$
  }
  \vspace{-1em}
  \label{fig:gaze_est}
\end{figure}

\begin{figure}[tb]
  \centering
  \includegraphics[width=85mm]{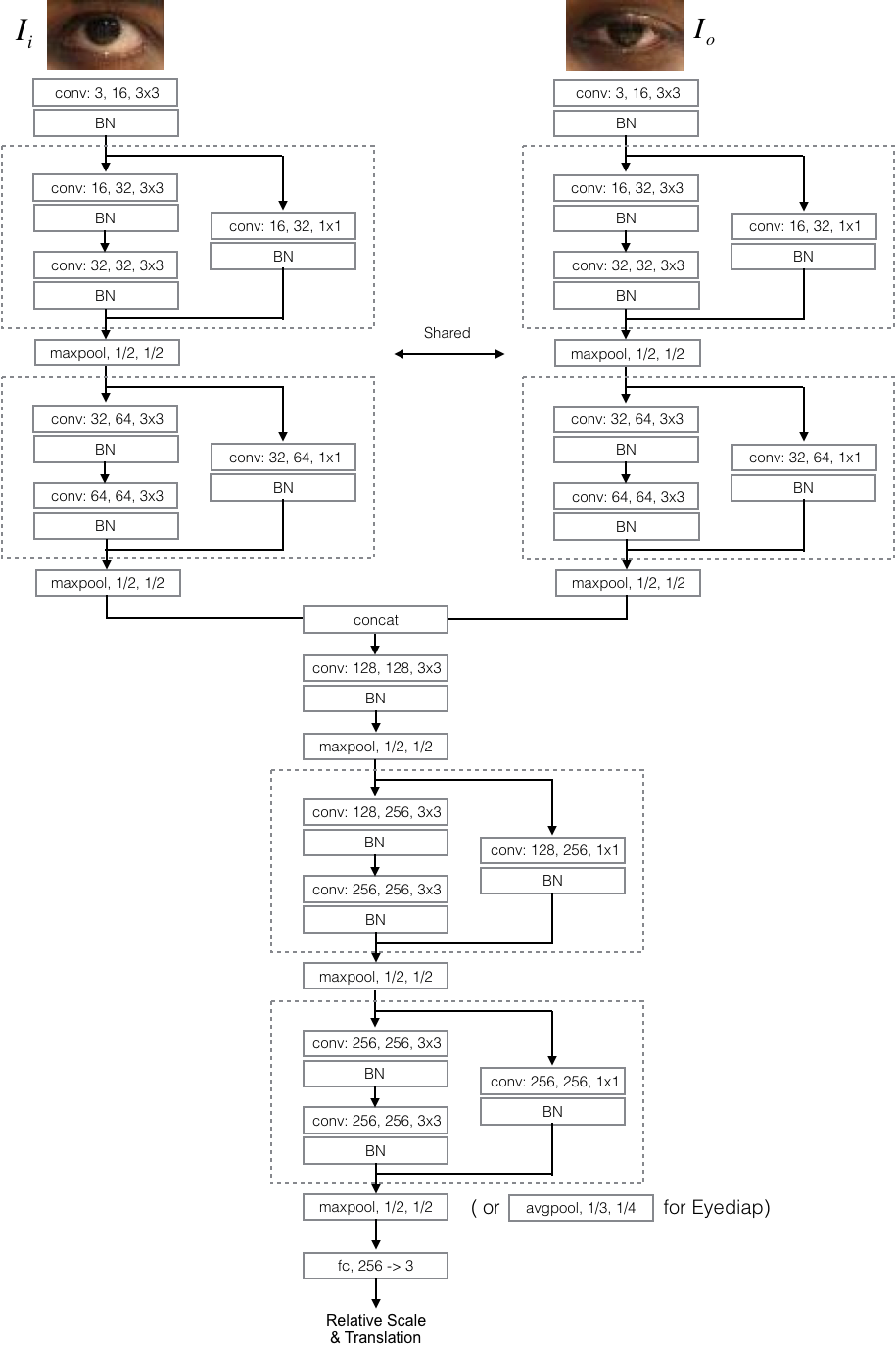}
  \caption{Global Alignment Network $\LNet$
  }
  \vspace{-1em}
  \label{fig:global_align}
\end{figure}

\begin{figure}[tb]
  \centering
  \includegraphics[width=85mm]{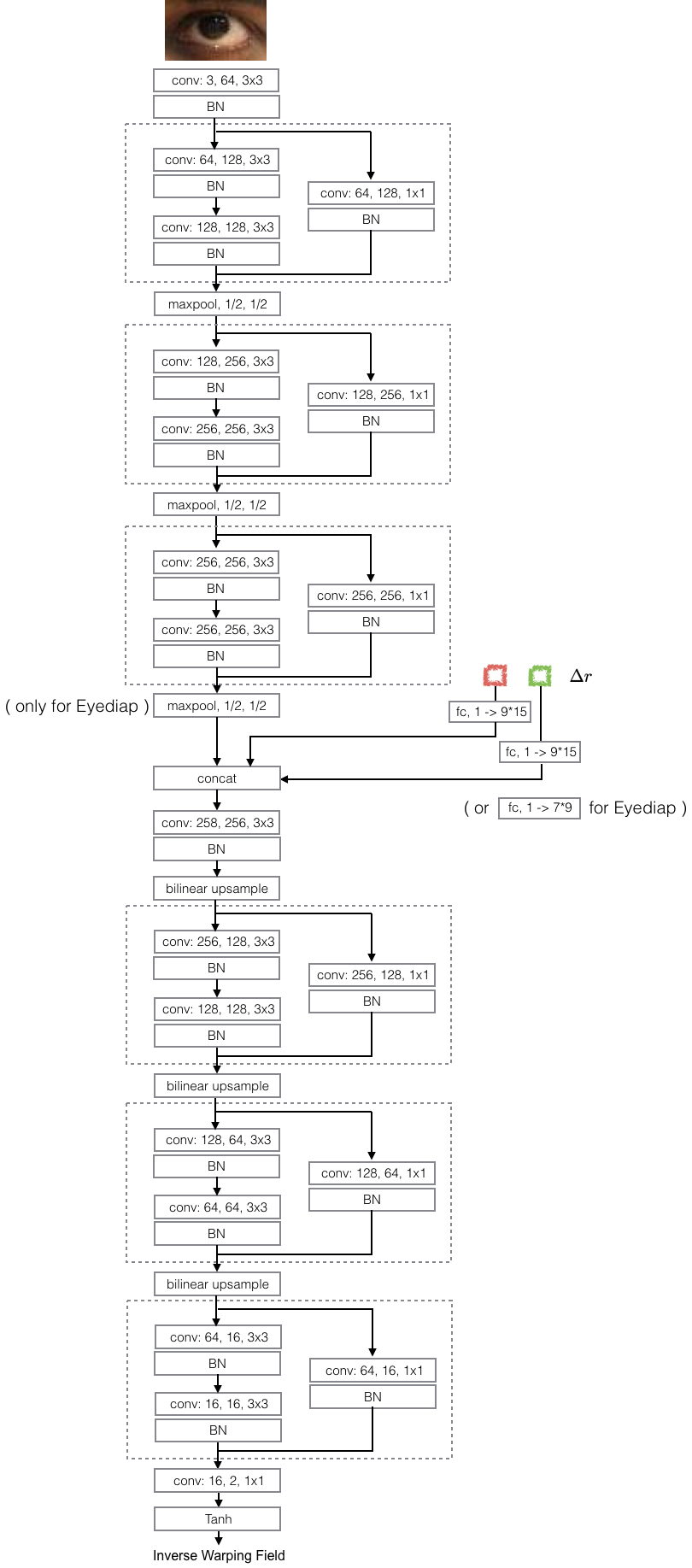}
  \caption{Gaze Redirection Network $\RNet$
  }
  \vspace{-1em}
  \label{fig:gaze_red}
\end{figure}

The detailed architectures of the Gaze Representation Learning Network $\GNet$,
the Global Alignment Network $\LNet$ and the Gaze Redirection Network $\RNet$
are illustrated in Fig.~\ref{fig:gaze_est}, Fig.~\ref{fig:global_align} and Fig.~\ref{fig:gaze_red}
respectively. They are based on ResNet blocks.
Because of the different image resolution (Eyediap input images are larger),
the architectures employed to handle
 Eyediap samples are a bit different than  those
for the Columbia Gaze and UTMultiview datasets.
They mainly differ in pooling operations, and have been
mentioned in the figures. 
Note that there are ReLu activation functions between the layers, which are omitted in the figures.

\subsection{Gaze Transfer}

\begin{figure}[tb]
  \centering
  \includegraphics[width=85mm]{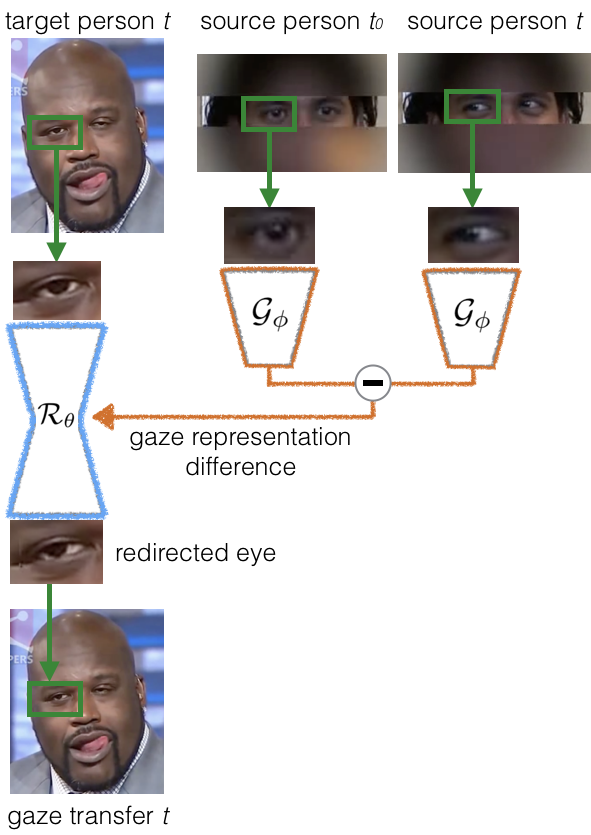}
  \caption{Gaze Transfer
  }
  \vspace{-1em}
  \label{fig:gaze_trans}
\end{figure}

With our proposed framework, we can also transfer the gaze movement of a source person to a target person (assuming no gaze movement for the target person).
More concretely, we extract the gaze representation of the eye of a source person at a time instant $t$ and a reference time $t_{0}$ via the gaze network $\GNet$. Then we compute the representation difference which encodes the gaze movement between time $t$ and $t_{0}$. At last, the gaze of the target person is redirected with the extracted representation difference
(hence is redirected towards the gaze of the source person).
In this way, the temporal gaze movement of the source person is transferred to the target person. Fig.~\ref{fig:gaze_trans} shows the whole procedure.
A demo video can also be found in: https://sites.google.com/view/yuyuvision/home.

Note that the learning of all network model parameters was done in a complete unsupervised fashion,
and at no point during training or for the transfer, gaze ground truth was needed.
%




\end{document}